\documentclass{article}
\usepackage{spconf,amsmath,graphicx}
\usepackage{booktabs}
\usepackage{multirow}

\title{PointResNet: Residual Network for 3D Point Cloud Segmentation and Classification}
\name{Author Name$^{\star \dagger}$ \qquad Author Name$^{\star}$ \qquad Author Name$^{\dagger}$}
\name{Aadesh Desai$^{*}$ \thanks{$^*$Equal Contribution} \qquad Saagar Parikh$^{*}$ \qquad Seema Kumari$^{*}$ \qquad Shanmuganathan Raman}
\address{Computer Vision, Imaging and Graphics Lab \\
	Indian Institute of Technology Gandhinagar, India\\
	\{desai.aadesh, saagar.p, seema.kumari, shanmuga\}@iitgn.ac.in}
\begin{document}
\ninept
\maketitle
\begin{abstract}
Point cloud segmentation and classification are some of the primary tasks in 3D computer vision with applications ranging from augmented reality to robotics. However, processing point clouds using deep learning-based algorithms is quite challenging due to the irregular point formats. Voxelization or 3D grid-based representation are different ways of applying deep neural networks to this problem. In this paper, we propose PointResNet, a residual block-based approach. Our model directly processes the 3D points, using a deep neural network for the segmentation and classification tasks. The main components of the architecture are: 1) residual blocks and 2) multi-layered perceptron (MLP). We show that it preserves profound features and structural information, which are useful for segmentation and classification tasks. The experimental evaluations demonstrate that the proposed model produces the best results for segmentation and comparable results for classification in comparison to the conventional baselines.

\end{abstract}
\section{INTRODUCTION}
\label{sec:intro}

3D point cloud has a wide range of applications in various areas such as robotics, animation, autonomous driving, industrial metrology, augmented reality, etc.~\cite{zhao2020point, Kumari2019AUTODEPTHSI}. Point cloud data represents 3D geometry as embedded sets of data points in continuous space, unlike images that are arranged on regular pixel grids. Hence, it makes the structural representation of 3D points different from the images. As a consequence, deep neural networks, such as convolutional neural networks (CNN), developed for 2D images can not be directly applied in processing 3D point clouds.


In order to address the issue mentioned above, various methods have been reported in the literature. Some methods voxelize the 3D points to employ the 3D discrete convolution~\cite{7353481,song2016ssc}. However, it increases the computational complexity and the memory requirement for processing the 3D points. Prior works like sparse convolutional neural networks can also be employed, but they operate only on nonempty voxels~\cite{inproceedings}. Further, some of the approaches directly operate on 3D points and extract information via soft max-pooling layers~\cite{pointnet,qi2017pointnet,9506398}. Another family of methods combines the information in the form of 3D points set into a graph for directive passing~\cite{dgcnn,li2021deepgcns_pami}.

Prior literature~\cite{segres} has shown that residual blocks-based approaches are employed as base networks by most state-of-the-art segmentation methods for 2D images. In this context, we propose a ResNet~\cite{xiang2021walk} inspired architecture that accepts unordered 3D points directly as input and produces per-point segment or class labels. The network is developed with multi-layer perceptron (MLP) layers and ResBlocks. MLP layers perform convolution and ResBlocks facilitate better information flow to produce relevant features for segmentation and classification tasks. Hence, the main contribution of our network is to employ ResBlock along with MLP layers to produce features that are well-suited for segmentation.

The rest of the paper is structured as follows, Section 2 discusses some part segmentation and object classification related work. Section 3 demonstrates the proposed approach of 3D point cloud segmentation and classification. Section 4 shows the experimental study of our model and compares qualitative and quantitative results with baseline approaches. Section 5 concludes our work with future directions.

\section{RELATED WORK}
\label{sec:related}

One way to apply deep learning based methods such as CNN on 3D points is to transform these onto volumetric grid~\cite{7353481,7900038,7298801,riegler2017octnet,wang2017cnn}.
Further, 3D point clouds can also be projected onto several image planes and thus mapped into a 2-dimensional representation. The final output is generated by applying 2D CNN on the representations, and feature fusion~\cite{qi2016volumetric,7410471}. However, these methods were inefficient in creating dense grids on projection planes due to the sparse nature of point clouds. These planes can significantly impact recognition ability, and occlusion in 3D can affect the accuracy.

In the study of 3D point clouds, there has been some recent focus on applying deep learning for tasks such as segmentation~\cite{pointnet,qi2017pointnet,liu}, classification~\cite{pointnet,qi2017pointnet,xiang2021walk,simonovsky2017dynamic}, denoising~\cite{pointcleannet,hermosilla}, etc. In this paper, we address the shape classification and segmentation problems. Here, we discuss the state-of-the-art methods related to these problems. 



\begin{figure*}[t]
    \centering
    \includegraphics[width=\textwidth]{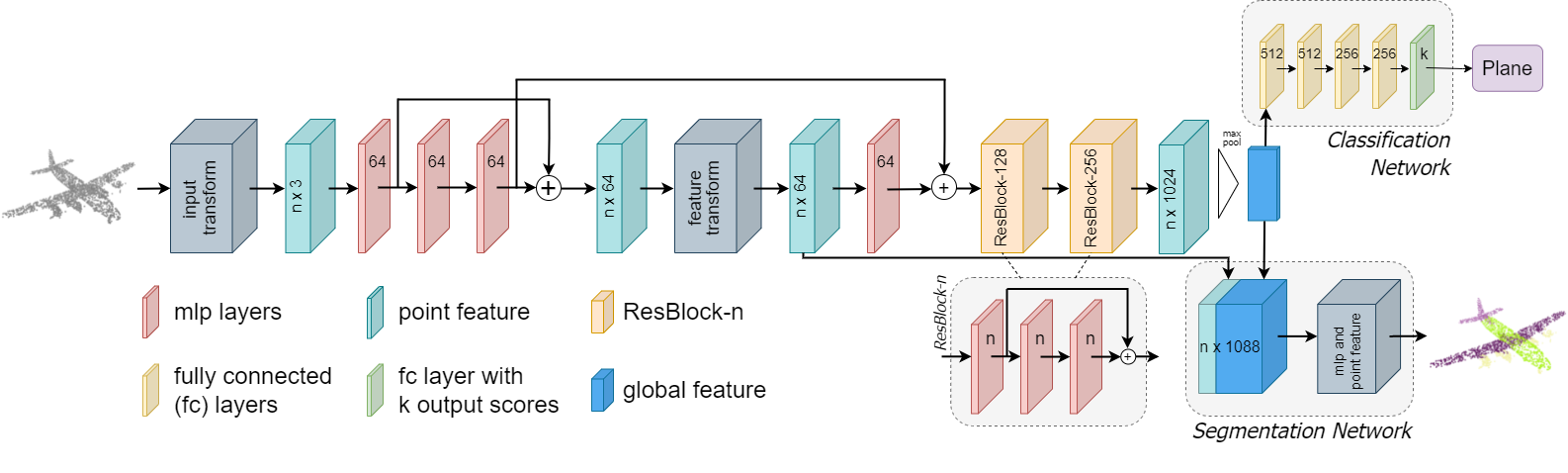}
    \caption{\textbf{PointResNet architecture} The network takes n input points and applies transformations, several multi-layer perceptron (MLP) layers with skip connections, and then aggregates features using max pooling. The classification network applies fully connected and dropout layers and obtains k scores for k classes. The segmentation network extends the classification network by combining local and global features, using convolutional layers, and finally giving per point scores as output. The connected skip layers represent ResBlock-n, where n is the number of features of a single block layer. Numbers in MLP layers represent a number of features.(Zoom in for best view)}
\end{figure*}

{PointNet}~\cite{pointnet} solves the problem of order invariance caused by simple deep neural network (DNN) models by directly processing 3D point clouds. The architecture consists of three main components: max-pooling layers as symmetry function, shared multi-layer perceptrons (MLPs), and a local-global information fusion module to classify and segment 3D point clouds. However, this network cannot describe shape. PointNet++~\cite{qi2017pointnet} first partitions points set into overlapping local regions and then produces higher-level features by extracting, grouping, and processing the local features. 

{PointConv}~\cite{wu2019pointconv} and {KPConv}~\cite{thomas2019kpconv} follow continuous convolution strategy without quantization. Basically, these methods directly apply convolution to the 3D point set. It is a density-weighted convolution that can fully estimate the 3D convolution on any unordered set of 3D points. The weight and density functions are estimated by MLP networks and kernel, respectively. The work in \cite{pointnet} cannot capture the local structure, therefore making it difficult to learn more fine-grained patterns and complex scenes. {CurveNet}~\cite{xiang2021walk} proposes that continuous descriptors (continuous sequences of point segments) are sufficient to estimate point cloud geometry.
{Rotation Transformation Network (RTN)}~\cite{deng2021rotation} attempts to extract pose information from 3D objects for analyzing point cloud under self-supervision.



In deep learning-based models, it is hard to ensure that the finally generated features are better than the intermediate features. ResNet~\cite{xiang2021walk} improves the scenario by including identity mapping. It enables the researchers to build deeper network architectures that can learn the difference between the input and target. It ensures better accuracy in the results. Proceeding in this direction, we propose to include ResBlocks along with MLP and CNN layers. The final features of this model are rich in information that enables better segmentation results while performing not worse than the state-of-the-art classification results.


\section{PROPOSED APPROACH}
A point cloud is defined as an unordered set of 3D points. Each 3D point $P_i$ is a vector that contains information about the $x, y,$ and $z$ coordinates of that point, along with extra details such as color and normal. These vectors of point clouds are directly used as the input to our approach.
\subsection{Network architecture}

In our proposed method, feature transformation network is similar to the PointNet~\cite{pointnet} model. PointNet can not capture local context at different scales, and hence by including different residual layers and additional mlp layers, we try to extract the missing information from the input data. While taking 3D point clouds as input, we need to consider the three main properties of 3D point clouds that differentiate them from 2D images: 1) unordered set, 2) neighboring points can form meaningful interactions, and 3) transformation invariance.

In this scenario, the network architecture (shown in Figure 1) is designed to consider these unordered point sets into the model directly. Further, we only require the cartesian coordinates of 3D points~\cite{pointnet}. Hence, we ignore the excess feature channels by mapping the unordered point set ${\{x_i\}}$ ${\in}$ ${R_d}$ to a vector $f_i$ by equation \ref{eq:1}.
\begin{eqnarray}
\label{eq:1}
f_i = g(\{h(\{x_i\}).
\end{eqnarray}
Here, the function $h$ depicts an MLP layer, and $g$ is a function of single variable along with maxpooling.
The transformed vector $f_i$ is passed through three MLP layers. MLP layer maps each $n$ input points to 64 dimensions. The introduction of residual layers in the MLP concatenates the input features to output features. This enables a smoother flow of information from the input to the output. 
\begin{eqnarray}
\mathcal{B}(y) = \mathcal{F}(y) + y.
\end{eqnarray}
The identity connection comes from the feature $y$. It learns the residual $\mathcal{F}(y)$. The $\mathcal{B}(y)$ is further transformed using Eq.~(\ref{eq:1}) to pass the relevant features. The output feature is then passed through an MLP layer, followed by a residual connection. The resultant features are then mapped to the dimension of 1024 via two ResBlocks. Each of the blocks consists of several MLP layers with a residual connection. The 1024-dimensional vector is fed to the segmentation network for the segmentation task. The same output vector is fed to the classification network for classification purposes.


A symmetric function (max pooling) is used to create a global feature vector to aggregate information from all the higher dimensional points. This vector ensures that the output is independent of the different input permutations. In the end, a five-layer fully-connected network assigns classification scores to the $n$ classes. 
For the segmentation task, we concatenate the local 64-dimensional feature with global features. In the end, we use MLPs to lower the dimension from 1088 to $m$ and give an output array of $n \times m$. Each row of the output matrix corresponds to each of the $n$ points, and the columns are related to the $m$ sub-categories. The local and global feature aggregation is quite essential to predict per-point quantities based on local and global-level semantics. Thus our architecture considers explicitly all the essential properties of the point cloud that require special attention.

Our architecture solves the problem of transformation invariance by introducing transformations such that output for a particular object remains the same even after rotational or translation transforms. The invariance problem is addressed by input and feature transformations sub-networks that normalize the pose for the input point cloud. The network applies transformations on the input point cloud similar to spatial transformers~\cite{spatial} to achieve pose normalization. A geometric transformation is equivalent to multiplying each point with a transformation matrix.

Thus during input transformation, a regression network predicts a $3 \times 3$ transformation matrix, which is multiplied by the $n\times3$ input. Similarly, feature transformation predicts a $64 \times 64$ transformation matrix, which gets multiplied with the $n\times 64$ input. 

The proposed architecture performs the segmentation and classification of 3D point clouds from a directly sampled point cloud. Our network adds a regularization term with the cross-entropy loss function to avoid over-fitting caused by the increased training parameters and instability during training. 

\begin{figure}[]
    \centering
    \includegraphics[width=0.4\textwidth]{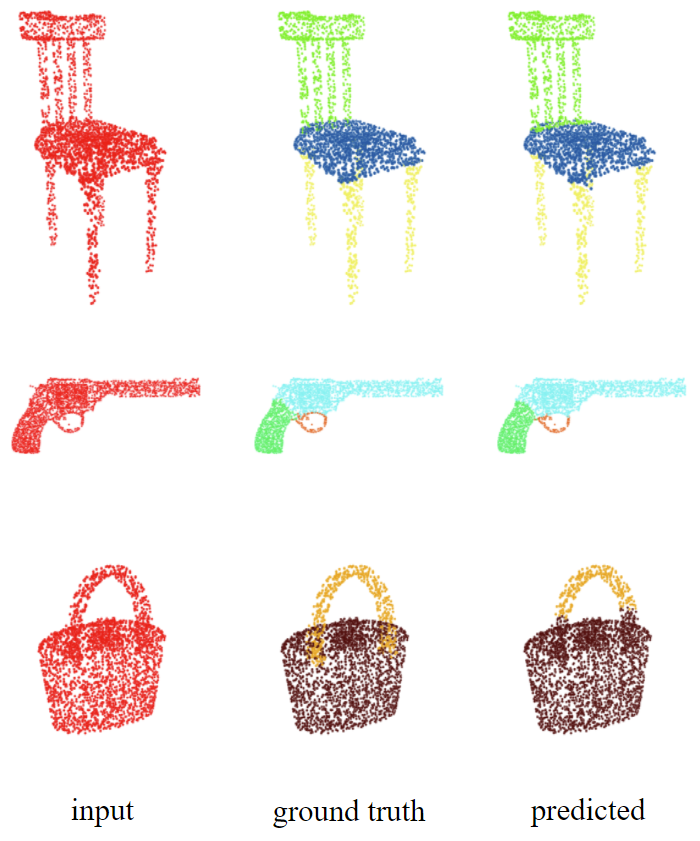}
    \caption{Visualization of part segmentation results on ShapeNetPart dataset.}
\end{figure}

\subsection{Loss function}
We use the softmax function on the output feature to calculate a probability score $\sigma(x_j)$ as shown in equation \ref{eq:softmax}.
\begin{eqnarray}
\label{eq:softmax}
\sigma(x_j) = \frac{e^{x_j}}{\Sigma_i e^{x_i}}
\end{eqnarray}
which is the used to compute the cross-entropy loss in equation \ref{eq:crossentropy}.
\begin{eqnarray}
\label{eq:crossentropy}
H(p, q) = - \Sigma_x p(x) \log q(x).
\end{eqnarray}
Here, \(p(x)\) = labels, and \(q(x) = \sigma(x_j)\) or \(1 - \sigma(x_j)\) depending on whether $j$ is the correct ground truth or not.

In addition to the cross-entropy loss, we include a regularization term to stabilize the training, which is given by equation \ref{eq:regul}.
\begin{eqnarray}
\label{eq:regul}
L_{reg} = ||I - AA^T ||^2.
\end{eqnarray}
Here, $A$ is the predicted feature alignment matrix. The feature transformation matrix is constrained to be closed to the orthogonal matrix. Hence, orthogonal transformation will have all the information of the input. Thus, the final loss function becomes as in equation \ref{eq:loss}.
\begin{eqnarray}
\label{eq:loss}
\text{Loss} = H + \alpha \times L_{reg},
\end{eqnarray}
where $\alpha$ is the assigned weight to the regularization term.

\section{EXPERIMENTS}


We evaluate our architecture on three standard datasets; ModelNet40~\cite{princeton}, ModelNet10~\cite{princeton}, and ShapeNetPart~\cite{team}. Further, we compare our results with the baselines and the existing methods in segmentation and classification tasks, which demonstrates the effectiveness of our model. The advantage of our architecture is further demonstrated by analyzing the effect of various parameters along with different architectures.


We implement our network on on a GeForce RTX 2080 Ti GPU using Tensorflow~\cite{tensorflow} environment. For training, we use 0.001 as learning rate that gets reduced with a decay parameter of 0.8. We choose 0.5 as the decay rate for batch normalization with a batch size of 32. For classification, we train our model for 250 epochs, and for segmentation, we train our model for 200 epochs with the adam optimizer.

\begin{table}[]
\begin{center}

\begin{tabular}{lc}
\hline
\multicolumn{1}{c}{}  & Eval acc \(\uparrow\)      \\ \hline
PointNet~\cite{pointnet}      & 90.6           \\ 
PointNet++~\cite{qi2017pointnet}    & 88.4           \\ 
ECC~\cite{simonovsky2017dynamic}      & 90.8           \\ 
RTN~\cite{deng2021rotation}           & 92.6           \\ \hline
PointResNet (Ours) & \textbf{94.79} \\ \hline
\end{tabular}
\caption{Quantitative comparison of part segmentation accuracy on ShapeNetPart dataset with the existing state-of-the-art methods.}\label{tab:seg}
\end{center}
\end{table}
\subsection{Datasets}

\textbf{ShapeNetPart}: ShapeNet is a large-scale dataset of 3D shapes containing 16 different single object classes. Each object includes many parts that can be segmented. 
\vspace{2mm}

\noindent \textbf{ModelNet}: The Princeton ModelNet project created the ModelNet40 dataset that includes 12,311 pre-aligned shapes from 40 categories. ModelNet10 is a subset of the ModelNet40 dataset that consists of 4,899 shapes from 10 categories. 

The train-test split for both datasets is 80\%-20\%.

\begin{figure}[]
    \centering
    \includegraphics[width=0.4\textwidth]{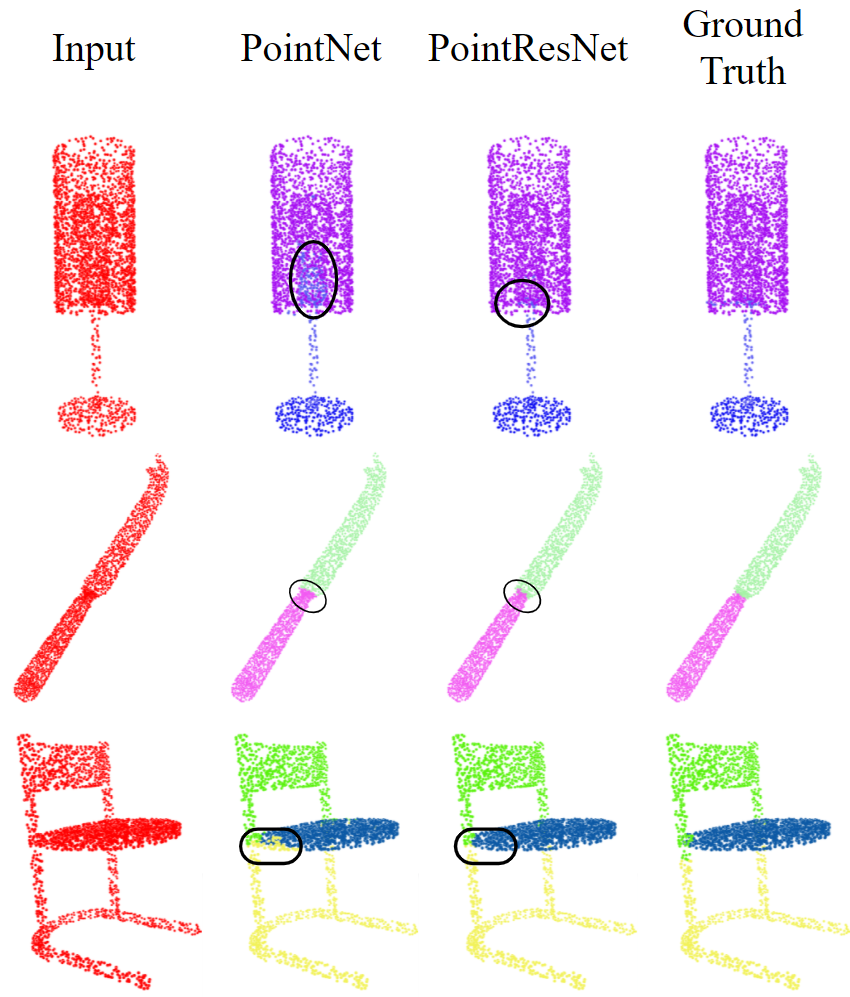}
    \caption{Qualitative comparison of part segmentation results on ShapeNetPart dataset.}
\end{figure}

\subsection{Results}

\subsubsection{Part segmentation on ShapeNetPart dataset}
Table \ref{tab:seg} demonstrates the quantitative comparison with our baselines and the state-of-the-art methods. It can be observed from Table \ref{tab:seg} that our proposed model outperforms the existing PointNet~\cite{pointnet}, PointNet++~\cite{qi2017pointnet}, ECC~\cite{simonovsky2017dynamic}, and RTN~\cite{deng2021rotation} methods for segmented parts. The reason for improvements can be observed visually in Figure 2, where segmented results produced by our model are very close to the ground-truths. Further, we also compare our results qualitatively with PointNet~\cite{pointnet} in Figure 3. One can note that our segmented part results are better than this method (see the marked area). Having achieved the desired results on segmentation using our proposed approach, we now validate our method for classification.



\subsubsection{Classification on ModelNet dataset}
Table \ref{tab:class} shows the comparison of our method with a representative set of the previously proposed state-of-the-art: PointNet~\cite{pointnet}, PointNet++~\cite{qi2017pointnet}, ECC~\cite{simonovsky2017dynamic}, RTN~\cite{deng2021rotation} and ResNet-50~\cite{perceiver}. We follow similar criteria to run our method on ModelNet datasets. To compare our results with these methods, we take the accuracy from the publicly available articles. In Table \ref{tab:class}, it can be noted that our proposed model produces better or comparable results for classification of the ModelNet datasets.


\begin{table}[]
\setlength{\tabcolsep}{1.5pt}
\begin{center}

\begin{tabular}{lcccc}
\hline
\multicolumn{1}{c}{} & \multicolumn{2}{c}{ModelNet10}                                                & \multicolumn{2}{c}{ModelNet40}                                               \\ \hline
\multicolumn{1}{c}{} & Eval acc \(\uparrow\)      & \begin{tabular}[c]{@{}c@{}}Eval avg\hspace*{2mm}\\ class acc \(\uparrow\) \end{tabular} & Eval acc \(\uparrow\)       & \begin{tabular}[c]{@{}c@{}}Eval avg\hspace*{2mm}\\ class acc \(\uparrow\) \end{tabular} \\ \hline
PointNet~\cite{pointnet}     & 92.52          & 92.08                                                        & 89.2          & 86.2                                                         \\
PointNet++~\cite{qi2017pointnet}   & -              & -                                                            & 91.9          & -                                                            \\
PointConv~\cite{wu2019pointconv}    & -              & -                                                            & \textbf{92.5} & -                                                            \\
ECC~\cite{simonovsky2017dynamic}         & 90.0           & -                                                            & 83.2          & -                                                            \\
RTN~\cite{deng2021rotation}          & -              & -                                                            & 90.2          & 86.5                                                         \\
ResNet-50~\cite{perceiver}          & -              & -                                                            & 66.3          & -                                                         \\
\hline
PointResNet (Ours)                 & \textbf{92.86} & \textbf{92.29}                                               & 88.76         & 85.58                                                        \\ \hline
\end{tabular}
\caption{Quantitative comparison of classification accuracy on ModelNet10 and ModelNet40 with the state-of-the-art methods.}    

\label{tab:class}
\end{center}
\end{table}

\subsection{Ablation study}
This section shows the importance of choosing 10 MLP layers and 4 residual blocks for our proposed model. For this task, we  implement three more architecture PointResNet 11, 15, and 16, as mentioned in Table 3. In PointResNet 11 model, we increase the sampling points from 1024 to 2048 with one MLP layer before transform-net.
Further, in comparison to PointResNet 11 model, the PointResNet 15 network consists of additional MLP layer of size 64,  three MLP layers of size 512, and one MLP layer of size 1024.

PointResNet 16 comprises of an encoder and decoder of UNet architecture~\cite{ronneberger2015u}.
Hence, we add 128 and 64 dimensional convolution transpose layers in the decoder part of the network. For segmentation task, PointResNet 11 and 16 models are evaluated on ShapeNetPart. The obtained accuracies are tabulated in Table\ref{tab:Ablation} (right column). Both accuracies (93.67\% and 94.73\%) are inferior than the proposed model's result.
In Table \ref{tab:Ablation} (second and third columns), we show the classification results on ModelNet 40 (third column) and ModelNet 10 (second column) datasets for various models. One can note that the performance of our PointResNet 10 is better as compared to other models.

\begin{table}[]
\begin{center}

\begin{tabular}{lccc}
\hline
\multicolumn{1}{c}{}                                            & \multicolumn{2}{c}{Classification}                                                                                            & \multicolumn{1}{c}{\begin{tabular}[c]{@{}c@{}}Part\\ Segmentation\end{tabular}}                                                \\ \hline
\multicolumn{1}{c}{}                                            & \begin{tabular}[c]{@{}c@{}}ModelNet10\\ Eval acc \(\uparrow\) \end{tabular} & \begin{tabular}[c]{@{}c@{}}ModelNet40\\ Eval acc \(\uparrow\) \end{tabular} & \multicolumn{1}{c}{\begin{tabular}[c]{@{}c@{}}ShapeNetPart\\ Eval acc \(\uparrow\) \end{tabular}} \\ \hline
PointResNet 15                                                  & 91.62                                                         & -                                                             & -                                                                                   \\
PointResNet 11                                                  & 92.29                                                         & 88.67                                                         & 93.67                                                                               \\
\begin{tabular}[c]{@{}l@{}}PointResNet 16\\ - UNet~\cite{ronneberger2015u}\end{tabular} & -                                                             & 88.75                                                         & 94.73                                                                               \\ \hline
PointResNet 10                                                  & 92.86                                                         & 88.76                                                         & 94.79                                                                               \\ \hline
\end{tabular}
\caption{Quantitative comparison of evaluated accuracy with proposed and our three newly implemented models by using the ModelNet40 and 10 dataset for classification and ShapeNetPart dataset for part segmentation.}
\label{tab:Ablation}  
\end{center}
\end{table}


\section{CONCLUSION}
We have proposed a deep neural network based on residual blocks and MLPs for the 3D point cloud segmentation and classification. Our proposed architecture directly operates on the 3D point cloud data instead of using any grid-based representation. The residual connection has shown an essential role in preserving the deeper feature information of 3D points essential for segmentation. The experimental results have depicted that our proposed unified framework is capable of performing outstanding segmentation and classifying as good as the baselines and state-of-the-art. 

We have implemented our network for part segmentation and object classification, and we plan to extend to segment semantic scenes by simply considering point labels as semantic object classes.  In future, we plan to improve the classification of 3D points using the data-augmentation technique. 


\bibliographystyle{IEEEbib}

\bibliography{references}

\end{document}